# EVLF-FM: Explainable Vision Language Foundation Model for Medicine

Yang Bai, PhD[1,#]

Haoran Cheng, MPH[2,3,#]

Yang Zhou, PhD[1,#]

Jun Zhou, PhD[1]

Arun Thirunavukarasu MA MB BChir[4,5]

Yuhe Ke, MBBS[6]

Jie Yao MD [2,3]

Kanae Fukutsu[2,3]

Chrystie Wan Ning Quek MBBS[3]

Ashley Hong, MBBS[3]

Laura Gutierrez MD [3]

Zhen Ling Teo FRCOphth [2,3]

Darren Shu Jeng Ting FRCOphth PhD [2,7,8,9,10]

Brian T Soetikno MD PhD [10]

Christopher S. Nielsen PhD [11]

Tobias Elze PhD [12]

Zengxiang Li PhD [3,13]

Linh Le Dinh PhD [1]

Hiok Hong Chan FRCOphth[3]

Victor Koh FRCOphth[14, 15]

Marcus Tan FRCOphth[14, 15]

Kelvin Z Li FRCOphth[16,17]

Leonard Yip FRCOphth[16,17]

Ching Yu Cheng MD PhD [3,15]

Yih Chung Tham PhD [3,15]

Gavin Siew Wei Tan MD PhD [2,3]

Leopold Schmetterer PhD [2,3]

Marcus Ang FRCOphth PhD [2,3]

Rahat Hussain MD FRCOphth [2,3]

Jod Mehta FRCOphth PhD [2,3]

Tin Aung MD PhD [2,3]

Lionel Tim-Ee Cheng FRCR [18]

Tran Nguyen Tuan Anh FRCR [18]

Chee Leong Cheng FRCPath [19]

Tien Yin Wong MD PhD[2,20,21,22]

Nan Liu PhD [23,24,25]

Iain Beehuat Tan FAMS PhD [2,26,27]
Soon Thye Lim FAMS PhD [2,26]
Eyal Klang MD[28]
Tony Kiat Hon Lim FRCPath[19]
Rick Siow Mong Goh PhD[1]
Yong Liu PhD[†1]
Daniel Shu Wei Ting MD PhD [†2,3,10,13]

1. Institute of High-Performance Computing (IHPC), Agency for Science, Technology and Research (A*STAR), Singapore
2. Duke-NUS Medical School, Singapore, Singapore
3. Singapore Eye Research Institute, Singapore National Eye Center, Singapore
4. Nuffield Department of Clinical Neurosciences, Medical Sciences Division, University of Oxford, Oxford, UK
5. International Centre for Eye Health, London School of Hygiene and Tropical Medicine, London, UK
6. Department of Anesthesia and Perioperative Medicine, Singapore General Hospital, Singapore
7. Department of Inflammation and Ageing, College of Medicine and Health, University of Birmingham, UK.
8. Birmingham and Midland Eye Centre, Sandwell and West Birmingham NHS Trust, Birmingham, UK.
9. Academic Ophthalmology, School of Medicine, University of Nottingham, Nottingham, UK.
10. Byers Eye Institute, Stanford University School of Medicine, Palo Alto, California.
11. Cumming School of Medicine, University of Calgary, Calgary, Alberta, Canada
12. Massachusetts Eye and Ear, Department of Ophthalmology, Harvard Medical School, USA
13. SingHealth AI Office, Singapore
14. Department of Ophthalmology, National University Health System, Singapore
15. Department of Ophthalmology, National University of Singapore
16. Department of Ophthalmology, National Healthcare Group, Singapore
17. Lee Kong Chien School of Medicine, Nanyang Technology University
18. Department of Diagnostic Radiology, Singapore General Hospital, Singapore
19. Department of Anatomical Pathology, Singapore General Hospital, Singapore
20. School of Clinical Medicine, Beijing Tsinghua Changgung Hospital, Tsinghua Medicine, Tsinghua University, Beijing, China

21. Beijing Visual Science and Translational Eye Research Institute, Beijing Tsinghua Changgung Hospital Eye Center, Tsinghua Medicine, Tsinghua University, Beijing, China
22. School of Biomedical Engineering, Tsinghua Medicine, Tsinghua University, Beijing, China
23. Centre for Quantitative Medicine, Duke-NUS Medical School, Singapore, Singapore.
24. Programme in Health Services and Systems Research, Duke-NUS Medical School, Singapore
25. NUS AI Institute, National University of Singapore, Singapore, Singapore
26. Division of Medical Oncology, National Cancer Centre Singapore, Singapore
27. Genome Institute of Singapore, Agency for Science, Technology and Research (A∗STAR), Singapore, Singapore
28. Generative AI Research Program, Mt Sinai, New York, USA

#These authors contributed equally to this work as joint first authors
†These authors contributed equally to this work as joint senior authors

**Corresponding Author:**
Assoc Prof. Daniel Ting MD PhD
Jong Soy Leong Professorship, Duke-NUS Medical School
Co-Director, AI Medicine Institute, SingHealth Duke-NUS Medical School
Singapore National Eye Centre, Singapore Eye Research Institute, Singapore
The Academia, 20 College Rd, Level 6 Discovery Tower, Singapore 169856
Email address: daniel.ting@duke-nus.edu.sg

**Financial Disclosure**: This study was supported by grants from the National Medical Research Council, Singapore (MOH-001689-00, MOH-000655-00 and MOH-001014-00), Duke-NUS Medical School (Duke-NUS/RSF/2021/0018, 05/FY2020/EX/15-A58, and 05/FY2022/EX/66-A128), the Agency for Science, Technology and Research, Singapore (A20H4g2141 and H20C6a0032), Research to Prevent Blindness and NIH P30 EY003790.

**Availability of Data**: Additional data may reasonably be requested from the corresponding author

**Conflict of interest:** DSWT holds a patent on a deep learning system for the detection of retinal diseases (10201706186V) and a computer-implemented method for training an image classifier using weakly annotated training data (10201901083Y) and stocks at EyRIS, Singapore and aISIGHT. AJT has received funding from HealthSense to support research concerning biomedical applications of large language models. The other authors declare no conflicts of interest.

**Abstract**

Despite the promise of foundation models in medical AI, current systems remain limited—they are modality-specific and lack transparent reasoning processes, hindering clinical adoption. To address this gap, we present EVLF-FM, a multimodal vision–language foundation model (VLM) designed to unify broad diagnostic capability with fine-grain explainability. The development and testing of EVLF-FM encompassed over 1.3 million total samples from 23 global datasets across eleven imaging modalities related to six clinical specialties: dermatology, hepatology, ophthalmology, pathology, pulmonology, and radiology. External validation employed 8,884 independent test samples from 10 additional datasets across five imaging modalities. Technically, EVLF-FM is developed to assist with multiple disease diagnosis and visual question answering with pixel-level visual grounding and reasoning capabilities. In internal validation for disease diagnostics, EVLF-FM achieved the highest average accuracy (0.858) and F1-score (0.797), outperforming leading generalist and specialist models. In medical visual grounding, EVLF-FM also achieved stellar performance in across nine modalities with average mIOU of 0.743 and Acc@0.5 of 0.837. External validations further confirmed strong zero-shot and few-shot performance, with competitive F1-scores despite a smaller model size. Through a hybrid training strategy combining supervised and visual reinforcement fine-tuning, EVLF-FM not only achieves state-of-the-art accuracy but also exhibits step-by-step reasoning, aligning outputs with visual evidence. EVLF-FM is an early multi-disease VLM model with explainability and reasoning capabilities that could advance adoption of and trust in foundation models for real-world clinical deployment.

**Introduction**

The emergence of foundation models (FMs) marks a paradigm shift in artificial intelligence (AI). Previously, AI generally referred to supervised deep learning (DL) models which were limited to diagnosing a narrow set of diseases from pre-specified labels.[1] In contrast, FMs are developed through large-scale unsupervised, semi-supervised, or self-supervised pretraining and fine-tuning that confers more general abilities such as simultaneous detection of multiple diseases, or accurate interpretation of images without specific training.[2–4] This versatility is compounded by a reduced requirement for labelled data (relative to conventional DL) to elicit useful performance. In addition, conversational AI systems extend FM capabilities by simulating clinician–patient interactions such as through structured history-taking and explanation of diagnosis and management.[5,6] These emerging applications illustrate the growing potential of FMs in medicine to support complex clinical reasoning and multimodal processing in diverse contexts, marking a significant step toward general medical AI (GMAI).[7]

However, FMs continue to exhibit important limitations. First, unimodal FMs, although highly performant within their own domains, are computationally expensive and impractical to scale across the breadth of clinical practice.[2,3] Multimodal models tend to underperform on tasks that require pixel-level precision.[8] These tasks include segmentation, where anatomical regions (*e.g.* organs, tumours, or vasculature) are annotated to facilitate measurement and comparison; as well as medical visual grounding, which involves defining relevant parts of an image in response to a query or report.[9,10] The trade-off between the multimodal versatility with pixel-level performance and portability raises concerns about whether these models can be used within high-stakes healthcare settings. A second persistent challenge is explainability: to foster trust, models should ideally demonstrate logical clinical reasoning in addition to generating accurate output. Although explainable AI (XAI) tools such as saliency maps and SHapley Additive exPlanations (SHAP) values have been explored in conventional DL systems, most are not directly applicable to emerging FMs.[11,12] Most FMs still function as black boxes, limiting transparency and reducing clinician confidence in their output.

To address these challenges, we introduce Explainable Vision-Language Foundation Model for Medicine (EVLF-FM), a state-of-the-art medical vision–language model (VLM) designed to integrate the breadth of generalist models with the pixel-level precision and explainability of specialist systems. EVLF-FM supports three core clinical tasks—disease diagnosis, visual grounding, and medical visual question answering (VQA)—while explicitly demonstrating its step-by-step reasoning. EVLF-FM was trained on one of the most diverse and comprehensive medical vision–language training corpora assembled to date, including over 1.3 million curated clinical images spanning six clinical specialties and ten imaging modalities such as computed tomography (CT), magnetic resonance imaging (MRI), X-ray (XR), ultrasound (US), positron emission tomography (PET), histopathology images, dermatoscopy, color fundus photography (CFP), optical coherence tomography (OCT), and endoscopic images. It incorporates supervised fine-tuning and visual reinforcement fine tuning within a generalist FM framework, aligning high-level diagnostic reasoning with fine-grained spatial localization. Unlike conventional models that provide only a final prediction, EVLF-FM generates intermediate justifications that mirror clinical reasoning, linking diagnostic outputs to visual evidence and clarifying how conclusions are reached. EVLF-FM enables both versatile performance across modalities and enhanced model interpretability, representing a towards the development of clinically trustworthy GMAI.

**Results**

To evaluate EVLF-FM's performance, we benchmarked the model against state-of-the-art generalist VLMs and specialist models across core medical AI tasks, including image-based diagnosis, VQA, and medical visual grounding, using a wide range of imaging modalities. Next, we assessed zero-shot and few-shot capabilities on external datasets not seen during training, comparing the performance of ELVF-FM with other VLMs. In addition, we conducted evaluations of data efficiency and consistency on repeat-prompting to test the reproducibility and robustness of model output.

**Internal validation across core medical AI tasks**

*Diagnostic performance in comparison to generalist VLMs and specialist DL models*
We benchmarked its performance in image diagnostics with four other generalist models (LLaVA v1.5 13B, Qwen VL 7B, LLava-Med V.5 7B, and InternVL 8B) and eight specialist DL models (VGG 16, ResNet 50, DenseNet 121, EfficientNet, ViT, DINO, CLIP, and SAM). Twelve datasets comprising over 1.3 million images were used for internal validation: PathMNIST, ChestMNIST, DermaMNIST, OCTMNIST, PneumoniaMNIST, RetinaMNIST, BreastMNIST, BloodMNIST, TissueMNIST, OrganAMNIST, OrganCMNIST, and OrganSMNIST.[13] Each of the generalist VLMs and specialist DL models is described in more detail in the Methods.

EVLF-FM demonstrated state-of-the-art diagnostic performance across the twelve test datasets, achieving a higher mean accuracy (0.858) and F1-score (0.797) than all of the comparator models (Supplementary Table 1). EVLF-FM exhibited superior F1-

scores to all general VLMs (Figure 2B) across every test dataset, and was also superior to all specialist DL models (Figure 2D) across the datasets with the exception of ViT for classification of OrganAMNIST data. In terms of accuracy, EVLF-FM outperformed general VLMs in eleven of twelve tasks (Figure 2A) and specialist DL models in ten of twelve tasks (Figure 2C). The exceptions were Chest-MNIST with LLaVA v1.5 13B and vgg16, and densenet exhibiting higher accuracy but inferior F1-score to EVLF-FM, as well as OrganAMNIST with ViT scoring marginally higher than EVLF-FM (Table 1). Of note, all models struggled with the diagnosis task for the ChestMNIST dataset, with accuracy ranging from 0.490 to 0.535. Nevertheless, EVLF-FM achieved a higher F1-score (0.253) than any other model tested on this dataset, indicating superior discriminative ability with an imbalanced dataset. Overall, where ELVF-FM exhibited consistently superior performance to all other models, and where it was surpassed by other models in two of twelve tasks, it ranked second and differences were slight.

In multiple validation tasks, the performance of EVLF-FM represented a qualitative advance on previous models (Supplementary Table 1). For instance, EVLF-FM achieved an accuracy and F1 score of 0.951 in the OCTMNIST dataset superior to four general VLMs with accuracy and F1 ranging from 0.738 to 0.868 and 0.729 to 0.868 respectively. Similarly, in TissueMNIST dataset, EVLF-FM attained an accuracy of 0.747 and an F1 score of 0.675, while other VLM models struggled, achieving accuracies between 0.575 and 0.642 (Figure 2A), and F1-scores from 0.411 to 0.540 (Figure 2B). Specialist DL models also struggled with accuracies between 0.456 to 0.678 (Figure 2C) and F1-scores between 0.215 and 0.593 (Figure 2D). Moreover, models with competitive or even superior scores to EVLF-FM in individual tasks exhibited dramatic drop-off when trialled with other datasets; an illustrative example being ViT which was the strongest model with OrganAMNIST, but much weaker than EVLF-FM with the DermaMNIST dataset. EVLF-FM exhibited superior or similar performance to VLMs and specialist DL models across the diverse range of tested modalities, with consistency not observed in any of the other models.

*Visual question answering (VQA)*
VQA tasks challenge models to generate accurate answers by integrating visual and textual information. We benchmarked EVLF-FM's performance in VQA against five specialised models: RadFM, LLaVA-Med, Med-Flamingo, MedDr, and InternVL 8B. Three datasets were used for internal validation: VQA-RAD, Slake-VQA, and Path-VQA. Model performance was evaluated in terms of closed accuracy with multiple-choice questions, as well as open recall with open-ended questions.

EVLF-FM demonstrated exceptional closed accuracy (Figure 2E) and open recall (Figure 2F). Closed accuracy was consistently high across all three datasets, with performance approaching 90% (Supplementary Table 1). Open recall, a more challenging metric that evaluates the model's ability to generate contextually appropriate responses rather than selecting from predefined options, typically presents greater difficulty due to the increased variability and complexity of free-text answers. Nevertheless, EVLF-FM achieved the highest open recall of all tested models on both VQA-RAD and SLAKE-VQA, exceeding 80%—a performance that highlighting the model's strong reasoning capabilities. In addition, EVLF-FM exhibited consistent performance across datasets and task formats. For example, while InternVL 8B achieved the highest open recall (77.5%) on the Slake-VQA dataset, its open recall dropped substantially to 52.9% for VQA-RAD dataset, suggesting limited consistency across different formats and topics. In contrast, EVLF-FM's consistently high performance in closed-recall accuracy demonstrates its precision in factual reasoning, while its strong open-ended recall reflects an ability to produce nuanced and contextually appropriate responses.

*Medical Visual Grounding*

We assessed EVLF-FM's medical visual grounding capabilities across nine different imaging modalities (MRI, pathology slides, XR, CT, OCT, CFP, US, dermatoscopy, and endoscopy) using conventional localization metrics: Acc@0.1, Acc@0.3, Acc@0.5 (accuracy at intersection-over-union thresholds of 0.1, 0.3, 0.5), and mean intersection over union (mIOU). In these evaluations, EVLF-FM was compared against a strong multimodal comparator (InternVL 8B). EVLF-FM consistently outperformed InternVL 8B across all metrics (Figure 2G). Averaged over all modalities, EVLF-FM achieved Acc@0.1 = 0.858, Acc@0.3 = 0.774, Acc@0.5 = 0.665, and mIOU = 0.626, surpassing the baseline models' corresponding scores of 0.844, 0.744, 0.613, and 0.577.

On a per-modality level, EVLF-FM showed particularly large gains for CT images (Accuracy@0.5 improved by over 11%) and endoscopic images (Accuracy@0.5 improved by over 14%), underscoring a robust ability to localize clinically relevant regions (Supplementary Table 2). For example, on endoscopic images, the mIOU increased from 0.608 with the baseline to 0.754 with EVLF-FM, indicating that the new model draws much tighter and more accurate bounding regions around actual anatomical targets. Although InternVL slightly outperformed EVLF-FM in a few isolated cases (*e.g.* pathology and X-ray at certain IoU thresholds), EVLF-FM's strong overall performance across a diverse set of image types was indicative of superior generalizability in visual grounding.

**External validation of zero-shot and few-shot performance**

*Comparison with VLMs for diagnostics*

We compared EVLF-FM's performance in image diagnostics against five other VLM models: Internvl3 8B, HuatuoGPT-Vision 7B, Meddr 32B MedGemma 4B, and MedGemma 27B. Six external dataset sets were used, spanning four different modalities (HRCTCov19 and SARS-COV-2 for CT, US3M for US, RetinalOCT for OCT, ATPOS and Glaucoma_fundus for CFP). For ELVF-FM, we evaluated both its zero-shot performance and few-shot performance, specifically 4-shot, 8-shot, 16-shot and 32-shot. Details about the dataset and VLM models can be found in method.

In the zero-shot setting, EVLF-FM achieved the highest F1-score and accuracy on three benchmark datasets: HRCTCov19, RetinalOCT, and Glaucoma_Fundus. (Figure 3A, Table 2) On the remaining three datasets, MedGemma-27B outperformed EVLF-FM with a narrow performance gap within 0.1 for both metrics. Importantly, EVLF-FM has a significantly smaller parameter size (8B) compared to MedGemma (27B), highlighting its efficiency relative to larger models. Additionally, MedDR-32B, another substantially larger model, only outperforms EVLF-FM on two datasets (SARS-COV2 and US3M) in terms of accuracy, but not in F1-score. These findings underscore the competitiveness and architectural efficiency of EVLF-FM, particularly given its lighter-weight configuration.

As anticipated, performance across diagnostic tasks improved with the incorporation of few-shot examples. (Figure 3B, Supplementary Table 3). The most substantial gains were observed with the RetinalOCT dataset, where accuracy increased from 0.481 (zero-shot) to 0.697 (64-shot), and F1-score from 0.338 to 0.604. Improvement was observed in every tested dataset, with the smallest gains seen with the HRCTCov19 dataset: accuracy increasing from 0.954 to 0.995 and F1-score from 0.949 to 0.994. Interestingly, increased shot count did not uniformly translate to better performance. On the US3M dataset, for instance, the 16-shot model yields the highest performance (accuracy = 0.604, F1-score = 0.703), whereas the 64-shot model achieved slightly lower scores (accuracy = 0.596, F1-score = 0.595). These results suggest that while few-shot tuning enhances diagnostic accuracy, optimal performance is not strictly dependent on the number of examples provided.

*Comparison with VLMs for grounding*

We benchmarked its performance in grounding with BiRD, a multimodal foundation model dedicated to grounding on an external data Med-GRIT-270k, comprising of 8 different modalities including CT, MRI, XR, PET, endoscopy, dermatoscopy, CFP, and US.[14] More details about the datasets and VLM can be found in the Methods.

EVLF-FM achieved strong performance on the Med-GRIT-270k benchmark (Supplementary Figure 1, Table 2). In a zero-shot setting with no Med-GRIT fine-tuning, EVLF-FM attained an average VG accuracy (Acc@0.5) of 36.17%. With only ~1% of the training data used for few-shot fine-tuning via our GRPO reward optimization, the few-shot model improved to 54.08%, surpassing the prior state-of-the-art BiRD model (53.92%) that used the full training set. Notably, EVLF-FM few-shot outperformed BiRD on several modalities – XR (41.7% vs 37.5%), PET (56.5% vs 53.8%), and US (54.7% vs 46.0% Acc@0.5) – despite being tuned on only a 1% data fraction. This demonstrated remarkable generalizability and efficiency of EVLF-FM's reward-guided tuning in comparison to a fully supervised baseline provided with 270,000 examples.

**Data efficiency analysis**

To assess how the amount of data used for second-stage reinforcement fine-tuning (RFT) affects model performance, we conducted a data efficiency analysis by training EVLF-FM with 20% (EVLF-20), 40% (EVLF-40), 60% (EVLF-60), and 80% (EVLF-80) of the original RFT dataset. We then evaluated the models on internal validation sets across three tasks. This analysis provided insight into the sensitivity of EVLF-FM's performance to the amount of reinforcement data used, informing the scalability and efficiency of fine-tuning.

EVLF-FM exhibited robust performance even when trained with a limited proportion of reinforcement data, indicating high data efficiency. On diagnostic tasks, EVLF-20—trained with only 20% of the reinforcement data—achieved results closely aligned with the fully trained EVLF-FM model, with differences in accuracy and F1-scores consistently below 0.01 across datasets. (Figure 3C, Table 3) For VQA tasks, EVLF-FM achieved approximately 3% higher performance in both closed-set accuracy and open-set recall across three datasets compared to EVLF-20. (Figure 3D, Table 3) In medical visual grounding, the observed performance gap between EVLF-20 and EVLF-FM was even narrower, with differences in Acc@0.5 and mIOU metrics as small as 0.02. (Figure 3E, Table 3) These findings collectively suggest that a relatively small volume of data is necessary to attain peak model performance, underscoring the efficiency of the training schema of EVLF-FM.

**Repeat-prompt reproducibility evaluation**

To assess the reliability of EVLF-FM, we evaluated its response consistency after rephrasing each clinical diagnostic question using ten distinct prompt templates. This experiment was conducted on the internal validation set for diagnostic tasks. The full set of prompt templates is detailed in Supplementary Table 4.

EVLF-FM demonstrated highly consistent performance across all 12 datasets even in the face of linguistic variation in question phrasing (Supplementary Table 4). As expected, minor decreases in accuracy and F1 scores were observed due to prompt diversity, which reflects the variability inherent in real-world clinical interactions. However, the magnitude of these differences was minimal. These findings underscore the model's robustness to natural language variation—an essential characteristic for dependable deployment in clinical practice.

**Discussion**

EVLF-FM is a multimodal medical foundation model designed with additional emphasis on human-like reasoning to enhance explainability in clinical settings. After training with one of the largest and most diverse clinical datasets assembled, EVLF-MED performs three key tasks: disease classification, medical VQA, and visual grounding. It demonstrates state-of-the-art performance and consistency in both internal and external validation, outperforming existing generalist VLMs as well as specialist architectures. Through a unique training strategy combining supervised and reinforcement fine-tuning, EVLF-MED also exhibits robust reasoning capabilities, aligning outputs with visual evidence. These findings demonstrate EVLF-FM's potential as a scalable, versatile and yet highly explainable clinical decision support tool, supporting efforts to develop trustworthy, human-aligned AI for healthcare applications.

EVLF-FM introduces several key innovations and strengths. The three core tasks of EVLF-FM—disease diagnosis, pixel-level visual grounding, and open-ended visual VQA—lend themselves to a variety of downstream clinical applications. These tasks are designed to mirror clinical reasoning by human experts: identifying anatomical structures on imaging, recognizing pathological findings, and localizing them precisely with confidence. By incorporating explanation-driven outputs through both explicit step-wise reasoning and pixel-level grounding, EVLF-FM delivers operational interpretability and auditability far beyond traditional black-box models—an essential feature for clinical decision-making and trustworthiness.[15] Another major advance stems from EVLF-FM's hybrid training strategy, which combines foundation model pretraining with supervised fine-tuning and a novel visual reinforcement fine-tuning phase implemented via GRPO. Supervised fine-tuning helps the model adapt to domain-specific tasks and ensures strong baseline performance, while requiring less task-specific data and computational overhead compared to training multiple specialist models.[16] Visual reinforcement fine-tuning addresses key limitations of supervised schemata by encouraging the model to generate structured, explainable reasoning rather than relying on memorized answers.[17,18] This is likely the driving factor underlying superior performance across all three tasks, surpassing other VLMs. By leveraging verifiable, task-specific rewards for reasoning coherence and grounding accuracy, this method promotes generalization to unfamiliar clinical settings and enhances interpretability, making EVLF-FM both efficient and trustworthy for real-world deployment robust to differences in image format.

The design and capabilities of EVLF-FM lend themselves to further development of applications that can change clinical practice and education. First, versatility across a multitude of clinical imaging media can support healthcare delivery in a range of settings, across many specialties. EVLF-FM can serve as a backbone model for innumerable downstream tasks, as fine-tuning can be undertaken to optimize performance for a specific task or to cater to demographic groups or clinical environments. Crucially, fine-tuning may be undertaken locally to preserve privacy if using local data. The ability of EVLF-FM to highlight areas of abnormality may benefit clinicians by guiding procedures and prioritizing attention or triaging images requiring closer scrutiny. Triage applications could help clinician cope with an ever-increasing workload to ensure further investigation and interventions are delivered as soon as possible.[19] In terms of diagnostic assistance, EVLF-FM may support differential diagnosis and clinical decision-making by integrating image interpretation with visual grounding and explanations—which could be especially useful in lower income settings where access to expertise or image reporting is more limited.[20]

Beyond clinical care, EVLF-FM holds significant value for medical education. It enables both medical students and junior doctors to explore cases through interactive diagnosis, grounding, and VQA, with visual justifications that support self-directed learning and reinforce clinical reasoning skills.[21] Additionally, similar capabilities could be adapted for patient-facing education, helping individuals better understand their health conditions and care plans. Integration with chatbot applications could further improve patient communication by explaining diagnoses, prognosis, and treatment options.[8,22] A prior study conducted in China found that patients with diabetic retinopathy who received care from both primary care providers and large language models (LLMs) exhibited better self-management behaviors and higher adherence to follow-up treatment compared to those managed solely by primary care providers.[23] These improvements are likely attributable to the LLM's capacity to deliver personalized, data-informed recommendations and to facilitate structured communication that supports shared decision-making between patients and clinicians.

Despite its strengths, EVLF-FM has several limitations that warrant further mitigation work. Notably, its performance on VQA tasks is significantly lower than on diagnostic classification tasks, a trend consistently observed across all foundation models. This discrepancy may be attributed to the greater complexity and open-ended nature of VQA, or to the relatively smaller number of annotated VQA samples available for training. Beyond task-specific differences, concerns about model generalizability and bias also remain. Like other large-scale models, EVLF-FM is susceptible to inheriting biases from its training data, which may lead to reduced performance in underrepresented patient populations, clinical environments, or imaging modalities.[24,25] Although our training dataset is large by current standards, it still captures a limited portion of real-world diversity. If the open-source data used is not representative—for instance, if certain demographics, imaging devices, or rare pathologies are under-sampled—EVLF-FM's predictions in these contexts may be less reliable.[24,25] Addressing these limitations will require future efforts to curate more diverse, high-quality, and representative datasets to ensure broader clinical applicability and fairness. Future work should prioritize curating more diverse and representative training datasets to mitigate this issue.

Additionally, the deployment of an AI model for autonomous clinical work raises ethical concerns such as the risks to patient safety of over-reliance on AI.[26] The potential negative impacts of AI assistance where models make errors are serious, and further model development and domain-specific validation are essential to ensure that error rates are minimized.[27] Moreover, clinicians retaining ultimate responsibility for interpretation and decision-making may help maintain confidence and trust, and ongoing discussion with all relevant stakeholders will be integral to determining how AI systems are integrated into clinical practice.[28] For clinical interventions involving autonomous AI, rigorous validation in prospective clinical studies, transparency in model use, and clear communication of confidence and uncertainty predictions will be necessary to build trust and minimize the risk of errors leading to harm.

Cost is another critical consideration in the deployment of foundation models in clinical settings. The high computational demands of large-scale models—particularly those requiring GPU acceleration—may pose significant barriers for many hospitals and healthcare systems, especially in resource-limited environments.[29] While prior real-world studies involving deep learning suggest that human-in-the-loop implementations are often the most cost-effective and pragmatic, no pilot studies have yet directly compared the cost-effectiveness of different deployment strategies for foundation models in clinical practice.[30] This remains an important area for future research to support equitable and sustainable integration of AI into healthcare systems.

In future work, we plan to validate EVLF-FM on additional real-world datasets and potentially pilot it in clinical settings to assess its impact on diagnostic accuracy, workflow efficiency, and clinician trust. We also envision using EVLF-FM as a foundation backbone for downstream applications, with further fine-tuning on modality-specific datasets to enhance performance in targeted clinical tasks and improve adaptability across diverse healthcare environments.

**Methods**

EVLF-FM aims to provide a single, unified solution for classification, detection, and interpretative tasks in the medical domain. The framework comprises three main components: 1) Disease-level encoder: A vision encoder that captures global, high-level image features to support improved diagnostic reasoning. This encoder is fine-tuned on large-scale medical imaging data to embed expert-level disease context into its representations. 2) Pixel-level encoder: A vision encoder that retains dense, spatially detailed features for region-specific analysis. This component preserves fine-grained pixel-level information (*e.g.* anatomical structures or lesion details) for precise medical grounding and detailed image understanding. 3) Training strategy: A multi-stage training regimen that first performs supervised fine-tuning (using low-rank adaptation, LoRA[31]) to align and specialize a large language model (LLM) for medical tasks, and then applies reinforcement learning via group relative policy optimization (GRPO) to directly optimize performance and reasoning. This approach balances parameter efficiency, preservation of prior vision–language knowledge, and domain-specific adaptation.

By unifying disease-level and pixel-level features within the model's architecture, EVLF-FM can address a wide variety of tasks without sacrificing either global domain insight or localized detail.

*Disease-level exemplar prompting*

To capture clinically salient patterns at a high level, EVLF-FM employs a disease-level encoder denoted $f_{disease}(\cdot)$ which is fine-tuned on large-scale medical imaging data. Given an input image $I$, this encoder produces a latent embedding vector $e$ representing the image's global disease-related features: $e = f_{disease}(I)$.

A learnable vision-language connector $\phi_{disease}(\cdot)$ then projects this disease-level embedding into the latent space of the language model, yielding a compatible representation $\hat{e} = \phi_{disease}(e)$.

Through this mechanism, the model can incorporate high-level medical expertise relevant to disease diagnosis and classification tasks. In practice, $\hat{e}$ serves as an exemplar prompt that informs the LLM of the image's overall pathology context, ensuring that subsequent text generation attends to clinically significant features.

*Pixel-level exemplar prompting*

To complement the global features, EVLF-FM integrates a pixel-level encoder $f_{pixel}(\cdot)$ that preserves spatially detailed information for localized analysis. Given the same image $I$, the pixel-level encoder produces a high-resolution feature map $p \in R^{H \times W \times C}$, where $H \times W$ is the spatial resolution and $C$ is the number of channels in the feature map. Instead of spatially pooling these features into one global feature, we pass the full-resolution feature map through a second connector $\phi_{pixel}(\cdot)$ to obtain a projected feature map $\hat{p}$ in the LLM's embedding space: $\hat{p} = \phi_{pixel}(p)$.

By preserving the complete grid of image features, EVLF-FM can supply the language model with fine-grained visual information. This pixel-level prompt enables the framework to perform explicit region-level reasoning, making it well-suited for tasks that require identifying specific anatomical locations or lesions (*i.e.* medical visual grounding and detailed image explanation).

*Unified exemplar prompting objective*

EVLF-FM jointly leverages the disease-level and pixel-level representations during training through a unified learning objective. For each training sample consisting of an image $I$, an input instruction or question $T$, and the corresponding ground-truth response $R$, the model conditions on both the global feature $\hat{e}$ and the fine-grained feature $\hat{p}$ to generate the response. We train the model to maximize the likelihood of the correct response given these inputs. Formally, the training objective is to minimize the negative log-likelihood:

$$L = -\frac{1}{N}\sum_{i=1}^{N} \log p_\theta(R_i \mid I_i, T_i, \hat{e}_i, \hat{p}_i),$$

where $N$ is the number of training examples and $\theta$ represents the trainable parameters of the model (including the LLM and the connector networks). In this formulation, the disease-level encoder $f_{disease}(\cdot)$ and pixel-level encoder $f_{pixel}(\cdot)$ provide crucial fixed representations (after their initial fine-tuning), and the connectors $\phi_{disease}(\cdot)$ and $\phi_{pixel}(\cdot)$ adapt those representations into the language model's space. By optimizing $L$, EVLF-FM learns to generate the appropriate response $R$ (such as a diagnosis, answer, or description) given both the image and the associated prompts, effectively uniting high-level and low-level visual cues in one learning process.

*Supervised fine-tuning with LoRA (Stage 1)*

In the first training stage, we perform supervised fine-tuning to align the multimodal model with medical domain knowledge and instruction-following capabilities. To preserve the pre-trained knowledge of the foundation model and ensure parameter efficiency, we freeze the core vision–language model weights – including the vision encoders and the original LLM parameters – and introduce LoRA layers into the LLM. We allow only the newly introduced LoRA parameters in the LLM and the connector weights ($\phi_{disease}$ and $\phi_{pixel}$) to be updated during fine-tuning. This way, the model learns to interpret the medical exemplar prompts and generate appropriate outputs for medical tasks, without overfitting or distorting the base model's general language and vision-language capabilities.

*Reinforcement Learning with Group Relative Policy Optimization (Stage 2)*

After supervised fine-tuning, we further refine EVLF-FM using reinforcement learning to directly optimize task performance metrics and encourage correct reasoning behavior. We adopt GRPO, a recent variant of the proximal policy optimization algorithm that is particularly efficient for language model fine-tuning. In conventional PPO, a learned value function (critic) is used to estimate a baseline for advantage calculation, which adds significant complexity and memory overhead, and it can struggle to assign per-token values in language generation. GRPO abrogates the requirement for a separate critic network; instead, it computes advantages by comparing a group of sampled outputs for the same input and using their average reward as an implicit baseline. This group-based relative advantage estimation aligns well with the comparative nature of reward signals in our setting (since our reward functions, described below, compare each output against the correct answer). It also eliminates the variance introduced by an absolute value baseline and simplifies the reinforcement learning pipeline.

In each RL update, we start with the policy model from the supervised fine-tuning stage (treated as the initial policy $\pi_{\theta_{old}}$) and sample a set of $G$ candidate outputs $\{o_i\}_{i=1}^{G}$ from it for a given input (e.g. a particular question or case). We then compute a reward $r_i$ for each output $o_i$ based on task-specific criteria (described below). Let $\bar{r} = \frac{1}{G}\sum_{i=1}^{G} r_i$ be the mean reward for that group of outputs. We define the advantage of each output $i$ as $A_i = r_i - \bar{r}$, which we further normalize by the group's standard deviation to obtain $\hat{A}_i$ (for numerical stability). Intuitively, outputs that score higher than the group average receive a positive advantage, and those below the average get a negative advantage. The model is then updated with policy gradient steps that encourage higher probability for tokens in outputs with positive $\hat{A}_i$ and lower probability for tokens in outputs with negative $\hat{A}_i$. To ensure stable learning, we apply the PPO-style clipping mechanism on the policy update and include a Kullback–Leibler (KL) divergence penalty to keep the updated policy from drifting too far from the pre-trained/SFT policy. Formally, the GRPO objective for the policy network is adapted from the clipped PPO loss as follows:

$$L_{\text{GRPO}}(\theta) = -E\left[\min\left(r_t(\theta)\widehat{A_t},\ \text{clip}(r_t(\theta),\ 1-\epsilon,\ 1+\epsilon)\widehat{A_t}\right)\right] + \beta\, D_{\text{KL}}(\pi_\theta\ |\ \pi_{\text{old}}).$$

Here $r_t(\theta) = \frac{\pi_\theta(a_t|h_t)}{\pi_{\text{old}}(a_t|h_t)}$ is the probability ratio at each token $t$ between the current policy $\pi_\theta$ and the reference (previous) policy $\pi_{\text{old}}$, and $\widehat{A_t}$ is the normalized advantage assigned to that token (in practice derived from the sequence-level $A_i$ above). The hyperparameters $\epsilon$ and $\beta$ control the clipping range and the strength of the KL penalty, respectively. This objective encourages the policy to improve on outputs with above-average reward and avoid regressing on outputs with below-average reward, while regularizing the policy update to prevent large deviations. Notably, GRPO achieves this without training a separate value function, significantly reducing the complexity and memory overhead of RL training and making it feasible to fine-tune a large multimodal model with limited data and compute.

*Reward design*

We crafted verifiable reward functions for each task so that the model could be trained with feedback signals grounded in objective correctness, without requiring any human in the loop. In our RL fine-tuning, we used only a very small fraction of the data (~1% of the SFT training set, on the order of a few thousand cases), focusing on examples with reliable ground-truth annotations to compute rewards. The reward for a given output was computed as follows for the key tasks:

Disease classification: The model's output in this task is typically a diagnosis (sometimes accompanied by a short explanation). We assign a reward of +1 if the output contains the correct diagnosis in the expected format, and 0 otherwise. This encourages the model to not only be correct, but also to present the answer in a clear, standardized way for clinical utility.

Pathology detection and localization: In this task, the model's answer includes identification of a finding and an indicated location (*e.g.* the model might output something like: "Detected finding: lung nodule in the left upper lobe; Location: (x, y, width, height)"). The reward is the localization accuracy. Specifically, we compute the Intersection-over-Union (IoU) between the model's predicted region (*e.g.* a bounding box or segmentation mask) and the ground-truth region marked by experts; higher IoU yields a higher reward, and the reward is zero if IoU is below a low threshold (meaning the model pointed to the wrong location entirely).

Open-ended QA (free-form reasoning): For general visual question-answering tasks requiring free-form explanations or reasoning, we did not apply reinforcement learning with direct numeric rewards—automatically grading open-ended answers goes well beyond simple exact-match criteria. Instead, we used F1-score (a standard evaluation metric in VQA that measures token-level overlap as the harmonic mean of precision and recall) to assess output quality—for example, treating the answer and ground truth as bags of tokens to compute F1-score.

During GRPO training, the model generates multiple candidate answers per query and receives the above reward for each; the group-relative advantage formulation then guides the model to preferentially produce answers that maximize these rewards. Notably, this process does not require a human preference model or manual rubric – the reward for each output is computed automatically against the known correct answer or annotation, making the reinforcement learning procedure stable and reproducible. We ran the GRPO fine-tuning for only a small number of iterations on the selected subset of data (again emphasizing efficiency). Even with this limited exposure, the impact on EVLF-FM was significant: the model learned to avoid minor errors that it sometimes made after the supervised fine-tuning stage (*e.g.* cases where it would output an almost-correct diagnosis but with slightly off terminology, or omit a required detail), and it became more consistent in producing well-formatted, complete answers. This translates to improved predictive accuracy – for instance, we observed higher exact-match accuracy in diagnosis and better IoU scores in localization after RL fine-tuning – as well as enhanced clinical interpretability of the model's outputs. The interpretability gains are especially evident in the detection task, where the model (when asked) can now pinpoint the finding on the image with greater precision and appropriately calibrated confidence, essentially providing a visual explanation for its diagnosis. We also observed that the model's open-ended explanations (when it provides reasoning or descriptions) became more aligned with clinical reasoning after RL training, likely because the model learned to focus on the key image features that maximize the reward (which, by design, correspond to the ground-truth findings). This outcome is in line with reports in other domains that policy optimization on a small subset of high-quality data can yield outsized improvements. By utilizing GRPO on roughly the top 1% of our data, we efficiently fine-tuned EVLF-FM to not only be more correct in its answers but also to present those answers in a manner that is trusted and useful to clinicians – a crucial aspect for real-world deployment.

In summary, our method blends architectural innovations with a comprehensive training regimen to produce a model with strong clinical reasoning abilities. The dual-encoder design provides EVLF-FM with a rich visual foundation: the disease-level encoder ensures broad diagnostic awareness of the image, while the pixel-level encoder provides granular visual evidence. The supervised training approach establishes the model's medical knowledge and basic vision–language alignment, with LoRA enabling effective adaptation on top of a powerful foundation model without overfitting, and GPT-4–augmented rationales seeding the model with improved explainability. Finally, GRPO-based reinforcement learning directly optimizes the model's behavior toward correct and explainable outputs, honing the model's focus on decision-critical details. Each step is deliberately designed to bridge the gap between raw medical images and the expert reasoning of a physician. The result is a multimodal foundation model–EVLF-FM–that achieves high performance on diagnostic tasks while also demonstrating interpretable reasoning and alignment with clinical norms. In practice, EVLF-FM can serve as a reliable AI assistant in medicine, capable of both answering, "What is it?" with accuracy and explaining, "Why?" in human-understandable terms, thereby maximizing its real-world utility for clinical decision support.

**Clinical Datasets**

A cornerstone of EVLF-FM's performance is its comprehensive, multi-task training dataset, curated to capture the full spectrum of clinical imaging diversity. In total, EVLF-FM was trained on 1,326,315 samples spanning three key tasks—disease diagnosis, visual grounding, and VQA—across a wide range of imaging modalities. This rich corpus not only ensures broad clinical applicability but also provides the fine-grained annotations necessary for detailed image interpretation.

For the disease diagnosis task, our training set comprises 554,391 samples derived from 9 imaging modalities. These samples were sourced from established datasets including MedMNIST, BRSET, AIROGS, and VinDr CXR. The MedMNIST collection comprises twelve two-dimensional biomedical datasets. It includes PathMNIST, with 100,000 colorectal-cancer H&E image patches and a centre-held-out test split of 7,180 patches for nine-class tissue recognition; ChestMNIST, containing 112,120 frontal chest radiographs from 30,805 patients annotated for 14 thoracic pathologies in a multi-label setting; DermaMNIST, offering 10,015 dermatoscopic images covering seven common skin conditions; OCTMNIST, featuring 109,309 retinal OCT B-scans grouped into four macular disease classes; PneumoniaMNIST, providing 5,856 paediatric chest radiographs labelled as pneumonia or normal; RetinaMNIST, comprising 1,600 fundus photographs relabelled with a five-level diabetic-retinopathy severity scale; BreastMNIST, consisting of 780 breast-ultrasound frames simplified to benign or malignant binary labels; BloodMNIST, with 17,092 peripheral blood-smear crops spanning eight white-cell phenotypes; TissueMNIST, containing 236,386 human kidney-cortex cell images partitioned into eight tissue classes; and Organ-A, Organ-C and Organ-S MNIST, which together contribute 30 861 axial, coronal and sagittal CT slices derived from LiTS and covering 11 abdominal organs.

Each image is paired with expert-verified diagnostic labels, enabling EVLF-FM to learn robust, high-level features essential for accurate classification across diverse clinical contexts.[32] The visual grounding dataset consists of 415,378 samples collected from eleven modalities. The training images for visual grounding were obtained from datasets such as Biomedparse[33] and Vindr, with each image annotated with bounding boxes that delineate key anatomical structures and pathological regions. These spatial cues empower EVLF-FM to perform visual grounding, thereby enhancing the interpretability of its diagnostic outputs.

For the VQA task, our corpus includes 208,230 samples sourced from datasets including Path-VQA, VQA-RAD, SLAKE, and PMC-VQA. This dataset comprises paired visual inputs and clinician-curated question–answer sets, facilitating the model's ability to integrate visual and textual information to generate precise, context-aware responses.

By integrating these three complementary datasets, EVLF-FM's training corpus spans 1,326,315 samples across eleven imaging modalities. The diverse and richly annotated data empower the model to capture both global diagnostic cues and localized visual details, ensuring high performance in disease diagnosis, visual grounding, and VQA. This extensive multi-task training is pivotal to EVLF-FM's ability to deliver state-of-the-art performance and to open new clinical vistas in AI-driven medical imaging.

*External datasets*

To gauge EVLF-FM's clinical generalizability, we evaluated it on an extensive panel of external datasets that together span fundus photography, optical coherence tomography (OCT), computed tomography (CT), ultrasound, and visual grounding tasks, drawing from both public and proprietary sources. For fundus-based disease grading, we used the APTOS dataset, which provides 1,100 test images for diabetic retinopathy (DR) severity classification, and MESSIDOR, which supplies 1,748 images from 874 examinations for the same task.[34,35] We further assessed referable-glaucoma detection on the GF dataset (465 images). Besides, we also evaluate on private dataset, Singapore Epidemiology of Eye Diseases Study (SEED-1) for DR grading, which contributes 810 images.[36] OCT performance was measured on RetinalOCT-C8, a test cohort of 2,800 images encompassing eight diagnostic categories—age-related macular degeneration, choroidal neovascularization, central serous retinopathy, diabetic macular edema, diabetic retinopathy, drusen, macular hole, and healthy controls.[37] For CT-based COVID-19 classification, we employed HRCT-Cov19 (913 scans) and SARS-CoV-2 CT (744 scans). Ultrasound robustness was examined with US3M, whose held-out set contains 230 cases labeled as benign or malignant liver lesions. Finally, spatial-localization capability was benchmarked on the large-scale Med-GRIT-270K corpus, which supplies dense bounding-box annotations across anatomical structures and pathologies; we used its full test split without further subdivision to provide a stringent external assessment.

**References**

1. LeCun, Y., Bengio, Y. & Hinton, G. Deep learning. *Nature* **521**, 436–444 (2015).
2. Zhou, Y. *et al.* A foundation model for generalizable disease detection from retinal images. *Nature* 1–8 (2023) doi:10.1038/s41586-023-06555-x.
3. Chen, R. J. *et al.* Towards a general-purpose foundation model for computational pathology. *Nat Med* **30**, 850–862 (2024).
4. Tanno, R. *et al.* Collaboration between clinicians and vision–language models in radiology report generation. *Nat Med* **31**, 599–608 (2025).
5. Tu, T. *et al.* Towards conversational diagnostic artificial intelligence. *Nature* 1–9 (2025) doi:10.1038/s41586-025-08866-7.
6. Thirunavukarasu, A. J. *et al.* Large language models in medicine. *Nature Medicine* **29**, 1930–1940 (2023).
7. Moor, M. *et al.* Foundation models for generalist medical artificial intelligence. *Nature* **616**, 259–265 (2023).
8. Zhang, K. *et al.* A generalist vision–language foundation model for diverse biomedical tasks. *Nat Med* 1–13 (2024) doi:10.1038/s41591-024-03185-2.
9. Wang, R. *et al.* Medical image segmentation using deep learning: A survey. *IET Image Processing* **16**, 1243–1267 (2022).
10. Yang, H. *et al.* Multimodal self-supervised learning for lesion localization. Preprint at https://doi.org/10.48550/arXiv.2401.01524 (2024).
11. Qi, Z. *et al.* A deep learning system for myopia onset prediction and intervention effectiveness evaluation in children. *npj Digit. Med.* **7**, 206 (2024).
12. Foo, L. L. *et al.* Deep learning system to predict the 5-year risk of high myopia using fundus imaging in children. *NPJ Digit Med* **6**, 10 (2023).
13. Yang, J. *et al.* MedMNIST v2 - A large-scale lightweight benchmark for 2D and 3D biomedical image classification. *Sci Data* **10**, 41 (2023).
14. Huang, X. *et al.* A Refer-and-Ground Multimodal Large Language Model for Biomedicine. Preprint at https://doi.org/10.48550/ARXIV.2406.18146 (2024).
15. Rosenbacke, R., Melhus, Å., McKee, M. & Stuckler, D. How Explainable Artificial Intelligence Can Increase or Decrease Clinicians' Trust in AI Applications in Health

Care: Systematic Review. *JMIR AI* **3**, e53207 (2024).

16. He, S. *et al.* GSCo: Towards Generalizable AI in Medicine via Generalist-Specialist Collaboration. Preprint at https://doi.org/10.48550/ARXIV.2404.15127 (2024).
17. Liu, Z. *et al.* Visual-RFT: Visual Reinforcement Fine-Tuning. Preprint at https://doi.org/10.48550/ARXIV.2503.01785 (2025).
18. Pan, J. *et al.* MedVLM-R1: Incentivizing Medical Reasoning Capability of Vision-Language Models (VLMs) via Reinforcement Learning. Preprint at https://doi.org/10.48550/ARXIV.2502.19634 (2025).
19. Annarumma, M. *et al.* Automated Triaging of Adult Chest Radiographs with Deep Artificial Neural Networks. *Radiology* **291**, 196–202 (2019).
20. Tan, T. F. *et al.* Artificial intelligence and digital health in global eye health: opportunities and challenges. *The Lancet Global Health* **11**, e1432–e1443 (2023).
21. Ng, F. Y. C. *et al.* Artificial intelligence education: An evidence-based medicine approach for consumers, translators, and developers. *CR Med* **4**, 101230 (2023).
22. Huo, B. *et al.* Large Language Models for Chatbot Health Advice Studies: A Systematic Review. *JAMA Network Open* **8**, e2457879 (2025).
23. Li, J. *et al.* Integrated image-based deep learning and language models for primary diabetes care. *Nat Med* **30**, 2886–2896 (2024).
24. Yu, A. C., Mohajer, B. & Eng, J. External Validation of Deep Learning Algorithms for Radiologic Diagnosis: A Systematic Review. *Radiology: Artificial Intelligence* **4**, e210064 (2022).
25. Yang, Y. *et al.* Demographic bias of expert-level vision-language foundation models in medical imaging. *Science Advances* **11**, eadq0305 (2025).
26. Liu, M. *et al.* A translational perspective towards clinical AI fairness. *npj Digit. Med.* **6**, 172 (2023).
27. Yu, F. *et al.* Heterogeneity and predictors of the effects of AI assistance on radiologists. *Nat Med* **30**, 837–849 (2024).
28. Thirunavukarasu, A. J. Large language models will not replace healthcare professionals: curbing popular fears and hype. *J R Soc Med* **116**, 181–182 (2023).
29. Zeng, D., Qin, Y., Sheng, B. & Wong, T. Y. DeepSeek's "Low-Cost" Adoption Across China's Hospital Systems: Too Fast, Too Soon? *JAMA* **333**, 1866–1869 (2025).
30. Xie, Y. *et al.* Artificial intelligence for teleophthalmology-based diabetic retinopathy screening in a national programme: an economic analysis modelling study. *Lancet Digit Health* **2**, e240–e249 (2020).
31. Hu, E. J. *et al.* LoRA: Low-Rank Adaptation of Large Language Models. Preprint at https://doi.org/10.48550/arXiv.2106.09685 (2021).
32. Zhang, S. *et al.* A Multimodal Biomedical Foundation Model Trained from Fifteen Million Image–Text Pairs. *NEJM AI* **2**, AIoa2400640 (2025).
33. Zhao, T. *et al.* A foundation model for joint segmentation, detection and recognition

of biomedical objects across nine modalities. *Nat Methods* **22**, 166–176 (2025).

34. Decencière, E. *et al.* FEEDBACK ON A PUBLICLY DISTRIBUTED IMAGE DATABASE: THE MESSIDOR DATABASE. *Image Analysis and Stereology* **33**, 231–234 (2014).
35. APTOS-2019 dataset. (2024).
36. Ahn, J. M. *et al.* A deep learning model for the detection of both advanced and early glaucoma using fundus photography. *PLOS ONE* **13**, e0207982 (2018).
37. Obuli Sai Naren. Retinal OCT Image Classification - C8. Kaggle https://doi.org/10.34740/KAGGLE/DSV/2736749 (2021).

**Figure 1.** Overview of the EVLF-FM framework and capabilities.

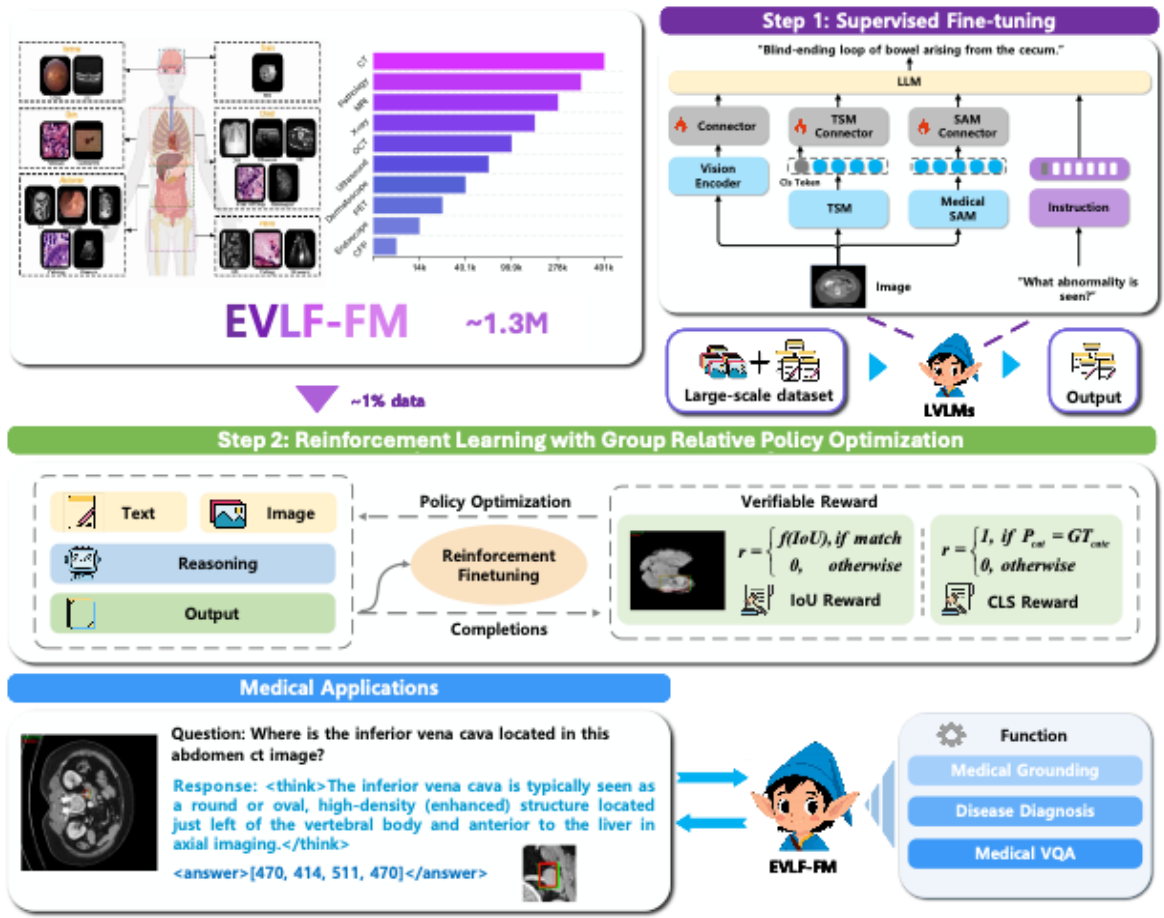

**Figure 2.** Internal validation results of EVLF, generalist vision-language models (VLMs), and specialist deep learning (DL) models in diagnosis across twelve datasets spanning a variety of modalities as well as in visual question answering (VQA). A) Accuracy results for EVLF-FM and VLM comparators across twelve diagnosis tasks. B) F1-score results for EVLF-FM and VLM comparators across twelve diagnosis tasks. C) Accuracy results for EVLF-FM and DL model comparators across twelve diagnosis tasks. D) F1-score results for EVLF-FM and DL model comparators across twelve diagnosis tasks. Overall, for diagnosis EVLF-FM represents a qualitative improvement over existing technology, with superior accuracy and F1-score in a large majority of tasks with competitive performance in the remainder (ChestMNIST and OrganAMNIST), exhibiting unprecedented consistency across modalities. E) Closed accuracy results for EVLF-FM and VLM comparators across three validation datasets. EVLF-FM exhibited superior closed accuracy to all comparator VLMs across the datasets. F) Open recall results for EVLF-FM and VLM comparators across three validation datasets. EVLF-FM exhibited superior performance to all comparator VLMs in two of three datasets, and was third to LLaVA-Med and MedDr in one dataset. G) Visual grounding results for ELVF-FM (right bars) and a VLM trained specifically for the task of visual grounding, BiRD (right bars). Evaluation was undertaken in terms of accuracy at insection-over-union thresholds of 10%, 30%, and 50% (Acc@0.1, Acc@0.3, and Acc@0.5) as well as mean instersection over union (mIOU). EVLF-FM exhibited superior performance to the VLM in every metric across all nine validation tasks.

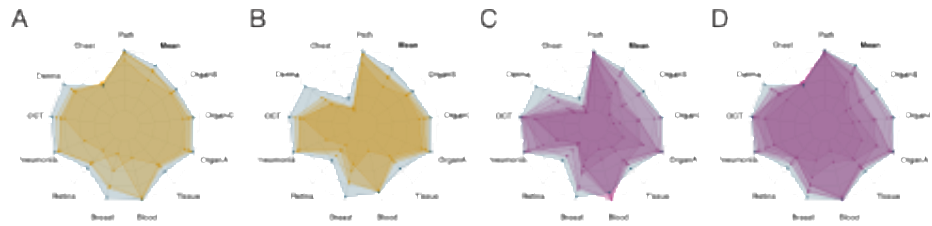
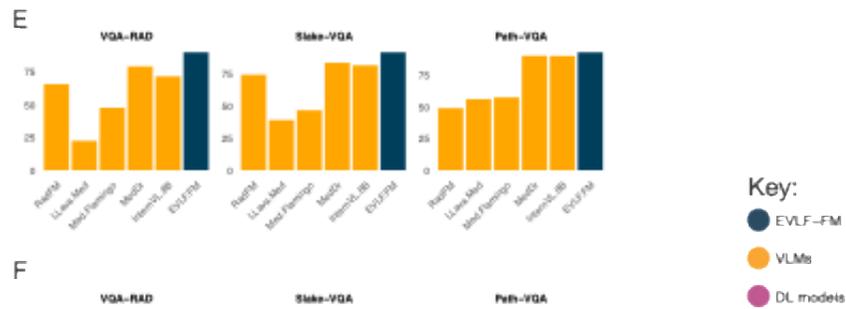
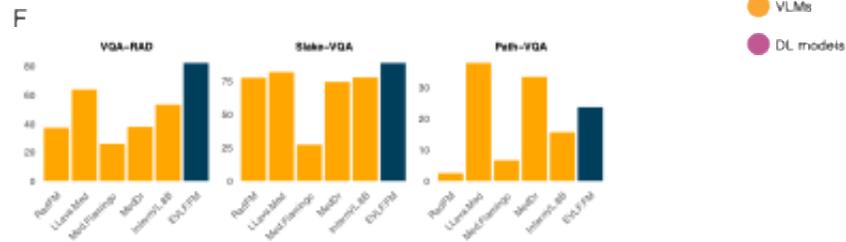
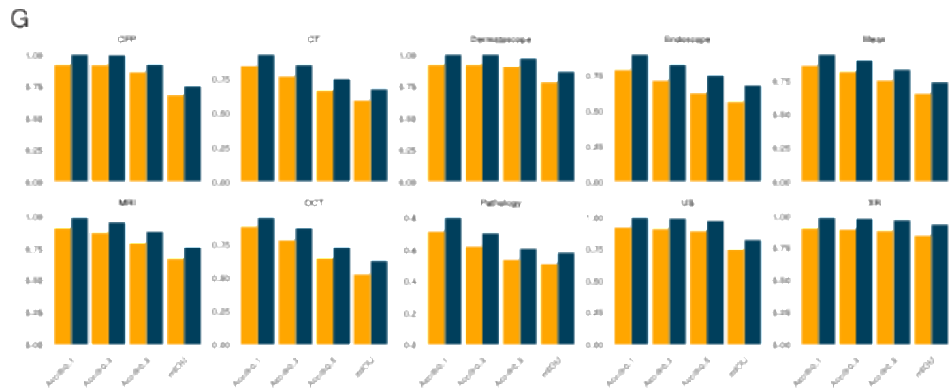

**Figure 3.** External validation performance of EVLF-FM in zero-shot and few-shot settings, tested against comparator vision-language models (VLMs) as well as with different pretraining and prompting schema applied. A) Zero-shot performance in terms of accuracy and F1-score of EVLF-FM and various other VLMs across six datasets spanning a wide range of clinical imaging modalities. EVLF-FM exhibited state-of-the-art performance for three benchmark datasets (HRCTCov19, RetinalOCT, and Glaucoma-Fundus) and was close to the top-performing model in the remaining three datasets; despite a much smaller size and computational requirement. B) Zero-shot and few-shot performance of EVLF-FM across the same diagnostic tasks, with improvement shown across most tasks as the number of examples provided of successful task completion increases. C) Diagnostic performance of EVLF-FM as the proportion of reinforcement fine-tuning data is adjusted from 20-100% in twelve tasks, with the mean performance coloured red. Only modest improvement is observed as the proportion of fine-tuning data increases, highlighting the data efficiency of EVLF-FM. D) Visual question-answering (VQA) accuracy and recall of EVLF-FM across three test datasets as the proportion of reinforcement fine-tuning data is adjusted from 20-100%. Clearer improvement is observed as fine-tuning data increases than seen for diagnostic performance. E) Medical visual grounding performance of EVLF-FM as the proportion of reinforcement fine-tuning data is adjusted from 20-100%. Modest but consistent improvement is observed as fine-tuning data increases, across all scoring metrics.

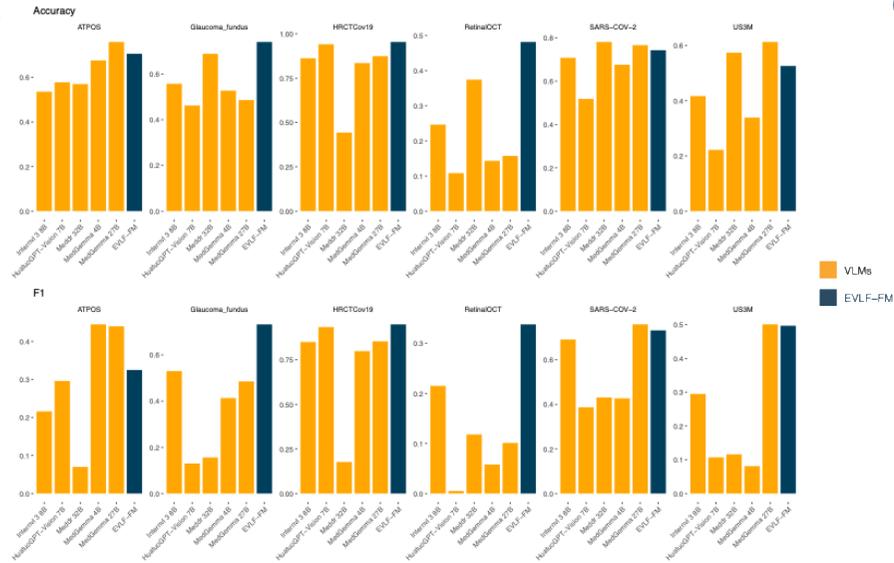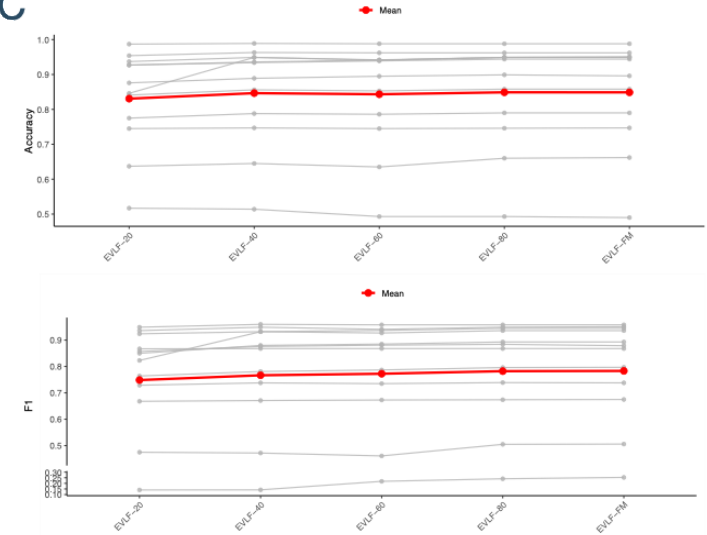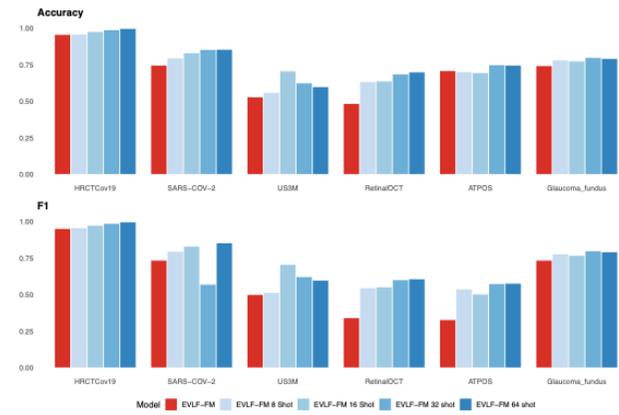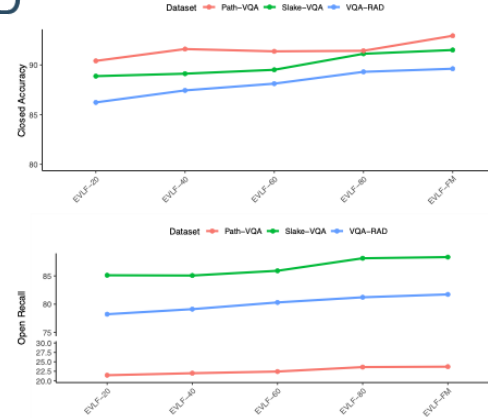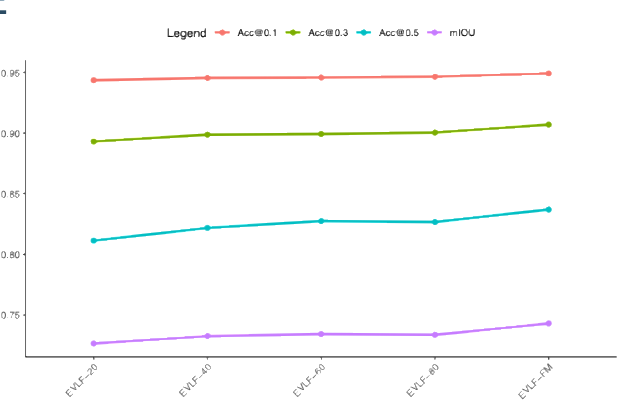

**Supplementary Figure 1**. Results on the performance of EVLF-FM for medical visual grounding in zero shot and few shot external validation against SOTA based on accuracy@0.5

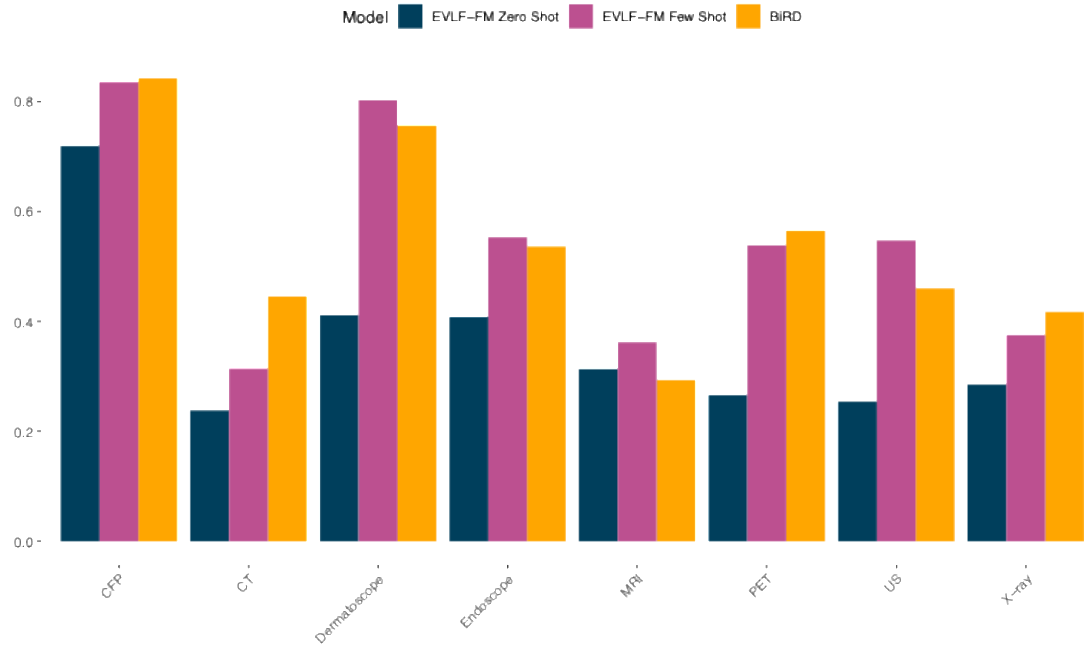

Table 1. Detailed results on the performance EVLF-FM for medical image disease diagnostics in internal validation against generalist VLMs and specialist DL models. Results are reported as accuracy/F1-score. Highest score for each evaluation matrix is highlighted in bold (including ties).

| | Path | Chest | Derma | OCT | Pneumonia | Retina | Breast | Blood | Tissue | Organ A | Organ C | Organ S | Mean |
|---|---|---|---|---|---|---|---|---|---|---|---|---|---|
| **Generalist Models** | | | | | | | | | | | | | |
| LLaVA v1.5 13B | 0.935/0.905 | **0.535**/0.073 | 0.731/0.355 | 0.788/0.786 | 0.881/0.864 | 0.557/0.279 | 0.750/0.671 | 0.951/0.832 | 0.613/0.497 | 0.878/0.855 | 0.796/0.750 | 0.689/0.621 | 0.759/0.624 |
| Qwen VL 7B | 0.823/0.754 | 0.510/0.051 | 0.716/0.384 | 0.738/0.729 | 0.438/0.383 | 0.280/0.166 | 0.494/0.510 | 0.286/0.166 | 0.575/0.411 | 0.807/0.777 | 0.724/0.681 | 0.672/0.618 | 0.589/0.469 |
| LLava-Med V1.5 (7B) | 0.939/0.915 | 0.513/0.088 | 0.800/0.556 | 0.868/0.868 | 0.910/0.900 | 0.542/0.280 | 0.212/0.382 | 0.975/0.856 | 0.642/0.540 | 0.916/0.908 | 0.865/0.843 | 0.738/0.687 | 0.743/0.652 |
| InternVL 8B | 0.938/0.911 | 0.520/0.045 | 0.755/0.487 | 0.798/0.788 | 0.886/0.872 | 0.575/0.394 | 0.788/0.699 | 0.960/0.841 | 0.594/0.465 | 0.865/0.838 | 0.817/0.783 | 0.698/0.650 | 0.766/0.648 |
| **EVLF-FM** | **0.969/0.960** | 0.490/**0.253** | **0.944/0.893** | **0.951/0.951** | **0.962/0.958** | **0.662/0.506** | **0.949/0.936** | **0.988**/0.868 | **0.747/0.675** | **0.949**/0.945 | **0.896/0.879** | **0.790/0.738** | **0.858/0.797** |

| Specialist Models | | | | | | | | | | | | | |
|---|---|---|---|---|---|---|---|---|---|---|---|---|---|
| vgg16 | 0.936/0.918 | 0.533/0.054 | 0.727/0.452 | 0.851/0.851 | 0.782/0.722 | 0.560/0.409 | 0.840/0.767 | 0.959/0.956 | 0.610/0.495 | 0.916/0.901 | 0.793/0.759 | 0.695/0.631 | 0.767/0.660 |
| resnet50 | 0.911/0.890 | 0.532/0.000 | 0.491/0.128 | 0.714/0.696 | 0.625/0.385 | 0.435/0.121 | 0.333/0.317 | 0.165/0.135 | 0.456/0.215 | 0.402/0.272 | 0.220/0.078 | 0.177/0.064 | 0.455/0.275 |
| densenet | 0.954/0.937 | 0.533/0.018 | 0.662/0.211 | 0.915/0.916 | 0.729/0.637 | 0.443/0.139 | 0.737/0.504 | 0.962/0.955 | 0.643/0.541 | 0.882/0.869 | 0.567/0.493 | 0.456/0.337 | 0.707/0.546 |
| efficientnet | 0.886/0.830 | 0.532/0.000 | 0.648/0.150 | 0.736/0.726 | 0.627/0.389 | 0.435/0.121 | 0.609/0.549 | 0.351/0.306 | 0.489/0.284 | 0.675/0.618 | 0.324/0.185 | 0.258/0.085 | 0.548/0.354 |
| VIT | 0.950/0.939 | 0.530/0.097 | 0.783/0.552 | 0.941/0.942 | 0.942/0.937 | 0.510/0.274 | 0.833/0.770 | 0.987/**0.988** | 0.673/0.583 | **0.952/0.950** | 0.851/0.835 | 0.732/0.674 | 0.807/0.712 |
| CLIP | 0.940/0.917 | 0.532/0.021 | 0.661/0.264 | 0.935/0.935 | 0.889/0.878 | 0.435/0.121 | 0.731/0.422 | 0.926/0.908 | 0.651/0.551 | 0.651/0.575 | 0.551/0.471 | 0.448/0.307 | 0.696/0.531 |
| DINO | 0.947/0.927 | 0.532/0.115 | 0.813/0.632 | 0.915/0.916 | 0.954/0.949 | 0.570/0.393 | 0.859/0.809 | 0.987/0.985 | 0.678/0.593 | 0.931/0.928 | 0.865/0.857 | 0.703/0.652 | 0.813/0.730 |
| SAM | 0.957/0.937 | 0.532/0.003 | 0.647/0.175 | 0.890/0.890 | 0.888/0.880 | 0.450/0.154 | 0.731/0.422 | 0.961/0.956 | 0.619/0.486 | 0.886/0.868 | 0.586/0.504 | 0.501/0.385 | 0.721/0.555 |

**Table 2.** Detailed results on the performance of EVLF-FM for medical image disease diagnostics and medical visual grounding in zero shot external validation against generalist VLMs. Results are reported as accuracy/F1-score. Highest score for each evaluation matrix is highlighted in bold (including ties).

| Diagnostics | CT | | US | OCT | CFP | |
| --- | --- | --- | --- | --- | --- | --- |
| | HRCTCov19 | SARS-COV-2 | US3M | RetinalOCT | ATPOS | Glaucoma_fundus |
| Internvl 3 8B | 0.862/0.850 | 0.708/0.691 | 0.417/0.295 | 0.246/0.215 | 0.536/0.216 | 0.557/0.530 |
| HuatuoGPT-Vision 7B | 0.941/0.934 | 0.518/0.387 | 0.222/0.107 | 0.108/0.005 | 0.578/0.296 | 0.462/0.130 |
| Meddr 32B | 0.443/0.177 | **0.781**/0.431 | 0.574/0.116 | 0.374/0.118 | 0.570/0.070 | 0.688/0.156 |
| MedGemma 4B | 0.835/0.799 | 0.676/0.427 | 0.339/0.081 | 0.143/0.058 | 0.676/0.445 | 0.527/0.413 |
| MedGemma 27B | 0.874/0.854 | 0.766/**0.759** | **0.613/0.501** | 0.157/0.101 | **0.759/0.440** | 0.486/0.485 |
| EVLF-FM | **0.954/0.949** | 0.743/0.732 | 0.526/0.497 | **0.481/0.338** | 0.706/0.325 | **0.740/0.732** |

| Grounding | CT | MRI | XR | PET | Endoscope | Dermatoscope | CFP | US | Mean |
| --- | --- | --- | --- | --- | --- | --- | --- | --- | --- |
| BiRD | **0.445** | 0.293 | **0.417** | **0.565** | 0.536 | 0.756 | **0.842** | 0.46 | 0.539 |
| EVLF-FM Zero Shot | 0.238 | 0.313 | 0.285 | 0.266 | 0.408 | 0.411 | 0.719 | 0.254 | 0.362 |
| EVLF-FM Few Shot | 0.314 | **0.362** | 0.375 | 0.538 | **0.553** | **0.802** | 0.835 | **0.547** | **0.541** |

Table 3. Detailed results on the performance of EVLF-FM for various tasks using varying proportions of RFT data in internal validation. Results are reported as accuracy/F1-score. Highest score for each evaluation matrix is highlighted in bold (including ties).

| Dataset | Evaluation Matrix | EVLF-20 | EVLF-40 | EVLF-60 | EVLF-80 | EVLF-FM |
|---|---|---|---|---|---|---|
| Diagnostics | | | | | | |
| **Path** | Accuracy/F1 | 0.965/0.955 | 0.966/0.957 | 0.968/0.958 | **0.969/0.960** | **0.969/0.960** |
| **Chest** | Accuracy/F1 | **0.517**/0.141 | 0.514/0.142 | 0.493/0.218 | 0.493/0.241 | 0.490/**0.253** |
| **Derma** | Accuracy/F1 | 0.927/0.850 | 0.934/0.880 | 0.939/0.885 | **0.944/0.893** | **0.944/0.893** |
| **OCT** | Accuracy/F1 | 0.937/0.936 | 0.949/0.949 | 0.941/0.941 | 0.949/0.949 | **0.951/0.951** |
| **Pneumonia** | Accuracy/F1 | 0.954/0.949 | **0.963/0.960** | 0.962/0.958 | 0.962/0.958 | 0.962/0.958 |
| **Retina** | Accuracy/F1 | 0.637/0.475 | 0.645/0.472 | 0.635/0.461 | 0.660/0.505 | **0.662/0.506** |
| **Breast** | Accuracy/F1 | 0.846/0.823 | 0.949/0.932 | 0.942/0.927 | **0.949/0.936** | **0.949/0.936** |
| **Blood** | Accuracy/F1 | 0.987/0.867 | **0.989/0.868** | **0.988/0.868** | **0.988/0.868** | **0.988/0.868** |
| **Tissue** | Accuracy/F1 | 0.745/0.668 | **0.747**/0.671 | 0.745/0.673 | 0.746/0.674 | **0.747/0.675** |
| **OrganA** | Accuracy/F1 | 0.927/0.924 | 0.936/0.932 | 0.941/0.937 | **0.949/0.945** | **0.949/0.945** |
| **OrganC** | Accuracy/F1 | 0.876/0.857 | 0.889/0.876 | 0.895/0.881 | **0.899/0.884** | 0.896/0.879 |
| **OrganS** | Accuracy/F1 | 0.775/0.729 | 0.788/0.738 | 0.786/0.735 | **0.790/0.739** | **0.790**/0.738 |
| Mean | Accuracy/F1 | 0.841/0.764 | 0.856/0.781 | 0.853/0.787 | **0.858**/0.796 | **0.858/0.797** |
| VQA | | | | | | |
| VQA-RAD | **Closed Accuracy** | 86.2 | 87.4 | 88.1 | 89.3 | **89.6** |
| | **Open recall** | 78.2 | 79.1 | 80.3 | 81.2 | **81.7** |
| Slake-VQA | **Closed Accuracy** | 88.9 | 89.1 | 89.5 | 91.1 | **91.5** |
| | **Open recall** | 85.1 | 85.1 | 85.9 | 88.1 | **88.3** |
| Path-VQA | **Closed Accuracy** | 90.4 | 91.6 | 91.4 | 91.4 | **92.9** |
|  | **Open recall** | 21.4 | 22 | 22.4 | 23.6 | **23.7** |
| Grounding |  | | | | | |
|  | Acc@0.1 | 0.944 | 0.946 | 0.946 | 0.947 | **0.949** |
|  | Acc@0.3 | 0.893 | 0.899 | 0.899 | 0.901 | **0.907** |
|  | Recall@0.5 | 0.81 | 0.822 | 0.828 | 0.827 | **0.837** |
|  | mIOU | 0.727 | 0.733 | 0.734 | 0.737 | **0.743** |

**Supplementary Table 1.** Detailed results on the performance of EVLF-FM for visual question-answering (VQA) tasks in internal validation against other VLMs. Highest score for each evaluation matrix is highlighted in bold (including ties).

|  | VQA-RAD | | Slake-VQA | | Path-VQA | |
|---|---|---|---|---|---|---|
|  | **Closed Accuracy** | **Open recall** | **Closed Accuracy** | **Open recall** | **Closed Accuracy** | **Open recall** |
| RadFM | 65.3 | 36.8 | 74.1 | 77.3 | 48.9 | 2.5 |
| LLava-Med | 22.3 | 63.4 | 38.9 | 81.6 | 56.0 | **37.9** |
| Med-Flamingo | 47.4 | 25.7 | 46.5 | 27.2 | 57.4 | 6.6 |
| MedDr | 78.9 | 37.5 | 83.4 | 74.2 | 90.2 | 33.5 |
| InternVL 8B | 71.3 | 52.9 | 81.4 | 77.5 | 90.1 | 15.7 |
| Ours | **89.6** | **81.7** | **91.5** | **88.3** | **92.9** | 23.7 |

**Supplementary Table 2.** Detailed results on the performance of EVLF-FM for medical visual grounding tasks in internal validation against other VLMs.

|  |  | MRI | Pathology | XR | CT | OCT | CFP | US | Dermatoscope | Endoscope | Mean |
|---|---|---|---|---|---|---|---|---|---|---|---|
| EVLF-FM | Acc@0.1 | 0.989 | 0.801 | 0.987 | 0.925 | 0.946 | 1.000 | 0.996 | 1.000 | 0.898 | 0.949 |
|  | Acc@0.3 | 0.954 | 0.703 | 0.978 | 0.850 | 0.869 | 0.994 | 0.989 | 1.000 | 0.827 | 0.907 |
|  | Acc@0.5 | 0.878 | 0.605 | 0.968 | 0.745 | 0.724 | 0.922 | 0.969 | 0.971 | 0.751 | 0.837 |
|  | mIOU | 0.758 | 0.581 | 0.934 | 0.673 | 0.625 | 0.752 | 0.821 | 0.866 | 0.679 | 0.743 |
| InternVL 3 8B | Acc@0.1 | 0.908 | 0.712 | 0.904 | 0.839 | 0.878 | 0.920 | 0.917 | 0.920 | 0.788 | 0.865 |
|  | Acc@0.3 | 0.871 | 0.618 | 0.893 | 0.766 | 0.775 | 0.918 | 0.907 | 0.920 | 0.713 | 0.820 |
|  | Acc@0.5 | 0.791 | 0.533 | 0.884 | 0.663 | 0.644 | 0.860 | 0.890 | 0.905 | 0.625 | 0.755 |
|  | mIOU | 0.670 | 0.506 | 0.846 | 0.589 | 0.523 | 0.682 | 0.746 | 0.783 | 0.564 | 0.656 |

**Supplementary Table 3.** Detailed results on the performance of EVLF-FM for medical image disease diagnostics in few shot external validation. Results are reported as accuracy/F1-score. Highest score for each evaluation matrix is highlighted in bold (including ties).

|  | CT | | US | OCT | CFP | |
| --- | --- | --- | --- | --- | --- | --- |
|  | HRCTCov19 | SARS-COV-2 | US3M | RetinalOCT | ATPOS | Glaucoma_fundus |
| EVLF-FM 0 Shot | 0.954/0.949 | 0.743/0.732 | 0.526/0.497 | 0.481/0.338 | 0.706/0.325 | 0.740/0.732 |
| EVLF-FM 8 Shot | 0.956/0.954 | 0.793/0.793 | 0.557/0.511 | 0.630/0.543 | 0.699/0.534 | 0.779/0.775 |
| EVLF-FM 16 Shot | 0.974/0.971 | 0.828/0.828 | **0.704/0.703** | 0.635/0.549 | 0.692/0.500 | 0.772/0.765 |
| EVLF-FM 32 shot | 0.986/0.984 | 0.851/0.567 | 0.622/0.619 | 0.683/0.598 | **0.746**/0.571 | **0.796/0.796** |
| EVLF-FM 64 shot | **0.995/0.994** | **0.852/0.851** | 0.596/0.595 | **0.697/0.604** | 0.743/**0.574** | 0.789/0.789 |

**Supplementary Table 4.** Detailed results on the performance of EVLF-FM for medical image disease diagnostics tasks using different prompts in internal validation. Results are reported as accuracy/F1-score. Highest score for each evaluation matrix is highlighted in bold (including ties).

|  | Path | Chest | Derma | OCT | Pneumonia | Retina | Breast | Blood | Tissue | Organ A | Organ C | Organ S | Mean |
|---|---|---|---|---|---|---|---|---|---|---|---|---|---|
| Base Prompt | **0.969/0.960** | **0.490/0.253** | **0.944/0.893** | **0.951/0.951** | **0.962/0.958** | **0.662/0.506** | **0.949/0.936** | **0.988/0.868** | **0.747/0.675** | **0.949/0.945** | **0.896/0.879** | **0.790/0.738** | **0.858/0.797** |
| Variant Prompt | **0.969/0.960** | 0.487/0.250 | 0.941/0.880 | 0.949/0.949 | 0.957/0.954 | **0.662/0.506** | 0.910/0.892 | **0.988/0.868** | 0.747/0.674 | 0.946/0.941 | **0.896/0.878** | 0.788/0.735 | 0.853/0.791 |

Base Prompt: Analyze the given {modality} image for diagnosis.

Variant Prompt 1: "Please perform diagnostic analysis on the provided {modality} image for diagnosis.

Variant Prompt 2: "Given a {modality} scan, determine the correct diagnosis"

Variant Prompt 3: "Evaluate the following {modality} image for diagnosis",

Variant Prompt 4: "Assess the {modality} image and provide a diagnosis",

Variant Prompt 5: "Based on the {modality} image, identify the diagnosis",

Variant Prompt 6: "For this {modality} image, specify its diagnosis",

Variant Prompt 7: "Analyze the provided {modality} scan for diagnosis",

Variant Prompt 8: "Interpret the {modality} image to determine its diagnosis",

Variant Prompt 9: "Diagnose the given {modality} image",

Variant Prompt 10: "Use the {modality} image to establish its diagnosis"